\def\year{2021}\relax
\documentclass[letterpaper]{article} % DO NOT CHANGE THIS
\usepackage{aaai21}  % DO NOT CHANGE THIS
\usepackage{times}  % DO NOT CHANGE THIS
\usepackage{helvet} % DO NOT CHANGE THIS
\usepackage{courier}  % DO NOT CHANGE THIS
\usepackage[hyphens]{url}  % DO NOT CHANGE THIS
\usepackage{graphicx} % DO NOT CHANGE THIS
\usepackage{natbib}  % DO NOT CHANGE THIS AND DO NOT ADD ANY OPTIONS TO IT
\usepackage{caption} % DO NOT CHANGE THIS AND DO NOT ADD ANY OPTIONS TO IT
\usepackage{amsthm}
\usepackage{amsmath, amssymb}
\usepackage[utf8]{inputenc} % allow utf-8 input
\usepackage[T1]{fontenc}    % use 8-bit T1 fonts
\usepackage{url}            % simple URL typesetting
\usepackage{booktabs}       % professional-quality tables
\usepackage{amsfonts}       % blackboard math symbols
\usepackage{nicefrac}       % compact symbols for 1/2, etc.
\usepackage{lipsum}
\usepackage{mathtools}
\usepackage{subcaption}
\usepackage[switch]{lineno}
\usepackage{algorithmic}
\usepackage{algorithm}
\graphicspath{ {./images/} }

\title{SecDD: Efficient and Secure Method for Remotely Training Neural Networks}

\author{
    Ilia Sucholutsky,\textsuperscript{\rm 1} Matthias Schonlau
}
\affiliations{
    Department of Statistics and Actuarial Science\\
    University of Waterloo\\
    Waterloo, Ontario, Canada
    \textsuperscript{\rm 1}isucholu@uwaterloo.ca

}
\begin{document}
\nocopyright
%\linenumbers
\maketitle
\begin{abstract}
We leverage what are typically considered the worst qualities of deep learning algorithms -  high computational cost, requirement for large data, no explainability, high dependence on hyper-parameter choice, overfitting, and vulnerability to adversarial perturbations - in order to create a method for the secure and efficient training of remotely deployed neural networks over unsecured channels. 

\end{abstract}

\section{Introduction}
We consider the situation where a neural network must be trained using proprietary or confidential data, but only an unsecured channel is available for providing data to the network. We assume that any data transmitted over this channel can be accessed by other parties. Our objective is to transmit data that will train the target network to desired accuracy, but be unusable by other networks, and also not reveal any information through qualitative inspection. A second objective is to improve efficiency by minimizing the size of our transmission. To this end, we propose using dataset distillation, the process of representing the knowledge of a large dataset using a smaller number of synthetic samples \citep{wang2018dataset}, as a method for efficiently and securely training neural networks. Specifically, Soft-Label Dataset Distillation (SLDD) is an extension to the dataset distillation algorithm that achieves even better performance by also learning distillation labels along with the distillation images \citep{sucholutsky2019softlabel}.  We propose Secure Dataset Distillation (SecDD) as an extension of SLDD that intentionally overfits samples to a target network in order to create tiny privacy-preserving training sets that reduce transmission size by by several orders of magnitude. These synthetic samples can only be trained on by a network with the same architecture and random initialization as the target network. These synthetic training samples can also be designed to qualitatively not resemble real samples; even appearing to belong to completely unrelated datasets. 

In order to retrieve private information from the synthetic samples, an attacker would need to discover both the architecture and random initialization of the target network. To do so, an attacker would have to perform Neural Architecture Search (NAS) on the synthetic training set. Fortunately, NAS methods are extremely computationally intensive and generally very data-hungry~\citep{Strubell_2019}. In particular, NAS has been shown to be ineffective when using small distilled datasets as proxies for the full training set \citep{shleifer2019using}. In addition, the search space for the NAS algorithm grows rapidly as the size of the target network increases. If the target network contains unusual components, it may even be impossible for NAS to find it as the search space is often constrained to popular network components. A good analogy for this is the process for creating a strong password: having a long password with special characters greatly increases the search space making it difficult for a brute-force attack to succeed.

\begin{figure*}
\centering
\includegraphics[width=0.49\textwidth]{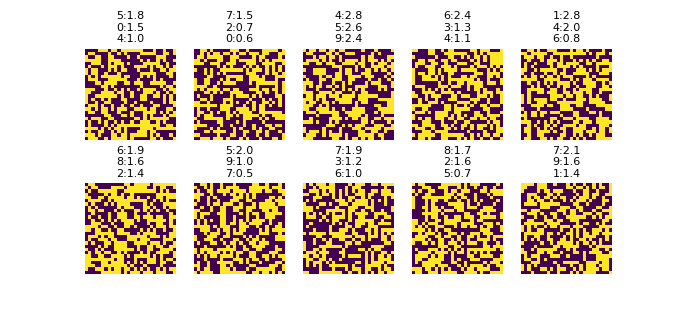}
\includegraphics[width=0.49\textwidth]{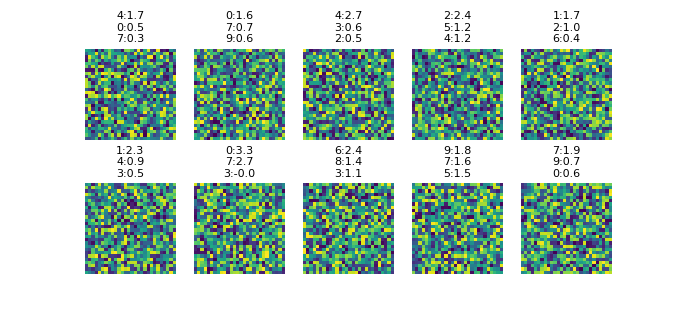}

\caption{SecDD can create various sets of 10 synthetic MNIST images that train target networks to over $95\%$ accuracy while visually appearing to consist almost entirely of noise. Each image is labeled with its top 3 classes and their associated logits.}
\label{fig:MNIST_knowninit}
\end{figure*}

\section{Related work}
% \subsection{Dataset distillation, data squashing, prototype selection, and prototype generation}
% Data squashing \citep{dumouchel2002data}

Prototypes have long been studied in the context of algorithms like k-nearest neighbours \citep{proto1, proto2}. Generally speaking, prototype methods aim to approximate datasets using a smaller number of samples. Prototype selection methods aim to choose prototypes from the actual dataset \citep{pselect2,pselect1}. Prototype generation methods, like the k-means algorithm, instead create synthetic samples \citep{pgen1, pgen2, pgen3}. Most prototype methods use hard labels, but some propose more complex prototypes that aim to increase efficiency \citep{mettes2019hyperspherical, sucholutsky2020oneshot}. Dataset distillation can be described as a family of prototype generation methods intended for use with neural networks \citep{wang2018dataset, sucholutsky2019softlabel, bohdal2020flexible}. Flexible Dataset Distillation (LD) is a recently proposed extension of dataset distillation that learns unrestricted labels as in SLDD but for a small fixed set of real images taken from the training dataset \cite{bohdal2020flexible}. 

% \citet{hu2020adversarial} proposed Adversarial Data Encryption as a method for modifying data such that it still looks the same for human observers, but misleads machine learning models. With SecDD, we instead aim to modify data such that it misleads both humans and machine learning models.

\section{Secure Dataset Distillation}
When using DD, and especially SLDD, with fixed initialization, the resulting distilled images qualitatively look mostly like random noise, yet they train the target network to impressive accuracies. Several studies criticized this behavior and proposed algorithms that result in clearer patterns in distilled images \citep{lorraine2019optimizing,zhao2020dataset,bohdal2020flexible}. However, we instead utilize the lack of interpretability as a way of preserving privacy by transmitting samples that do not resemble the ones in the original dataset. We modify the SLDD algorithm to encourage aggressive overfitting to the target network. While SLDD generally uses one-hot encoding to initialize the distilled labels, we experiment with alternative initializations that encourage more class mixing, and result in less identifiable features in the resulting distilled images. 
Flexible Dataset Distillation  uses fixed distilled images and instead only learns the associated soft labels. In fact, \citet{bohdal2020flexible} showed that the frozen images can come from a different dataset and still train the model to high accuracies on the target dataset. We propose two SecDD modes that leverage this idea in order to mask transmissions. In the first mode, fixed distilled images are initialized as random noise to ensure that attackers would not be able to discern qualitative features by observing transmissions. In the second mode, fixed distilled images are initialized as images taken from a different, completely unrelated dataset. For both modes, the soft labels for the images are learned through backpropagation. While aiding with privacy preservation, these modes may require larger distilled datasets to train models to the same accuracies than when using regular SLDD. Two example sets of fixed, random-noise samples used for training a target network to achieve high accuracy on MNIST are shown in Figure~\ref{fig:MNIST_knowninit}. The two sets used different initializations which resulted in visually different images, but both initializations still result in high accuracy for the target network.

\section{Conclusion and future work}
We have proposed a method for producing synthetic data that can be used to securely and efficiently train remotely deployed neural networks over unsecured channels. These transmissions can even appear to contain random noise or completely unrelated data while still training target neural networks to high accuracies. We have so far only conducted exploratory experiments to validate our claims and are working on conducting a comprehensive set of experiments that would quantify the improvements in privacy preservation and efficiency that SecDD can provide.

%\subsection{Figures}
%See Figure \ref{fig:fig1}. Here is how you add footnotes. \footnote{Sample of the first footnote.}

%\begin{figure}
%  \centering
%  \fbox{\rule[-.5cm]{4cm}{4cm} \rule[-.5cm]{4cm}{0cm}}
%  \caption{Sample figure caption.}
%  \label{fig:fig1}
%\end{figure}

%\subsection{Tables}
%See awesome Table~\ref{tab:table}.

%\begin{table}
% \caption{Sample table title}
%  \centering
% \begin{tabular}{lll}
%   \toprule
%   \multicolumn{2}{c}{Part}                   \\
%   \cmidrule(r){1-2}
%   Name     & Description     & Size ($\mu$m) \\
%   \midrule
%   Dendrite & Input terminal  & $\sim$100     \\
%    Axon     & Output terminal & $\sim$10      \\
%    Soma     & Cell body       & up to $10^6$  \\
%    \bottomrule
%  \end{tabular}
%  \label{tab:table}
%\end{table}

\bibliography{references}
\end{document}